# Utilizing Import Vector Machines to Identify Dangerous Pro-active Traffic Conditions*

Kui Yang, Wenjing Zhao, Constantinos Antoniou

*Abstract*— Traffic accidents have been a severe issue in metropolises with the development of traffic flow. This paper explores the theory and application of a recently developed machine learning technique, namely Import Vector Machines (IVMs), in real-time crash risk analysis, which is a hot topic to reduce traffic accidents. Historical crash data and corresponding traffic data from Shanghai Urban Expressway System were employed and matched. Traffic conditions are labelled as dangerous (i.e. probably leading to a crash) and safe (i.e. a normal traffic condition) based on 5-minute measurements of average speed, volume and occupancy. The IVM algorithm is trained to build the classifier and its performance is compared to the popular and successfully applied technique of Support Vector Machines (SVMs). The main findings indicate that IVMs could successfully be employed in real-time identification of dangerous pro-active traffic conditions. Furthermore, similar to the "support points" of the SVM, the IVM model uses only a fraction of the training data to index kernel basis functions, typically a much smaller fraction than the SVM, and its classification rates are similar to those of SVMs. This gives the IVM a computational advantage over the SVM, especially when the size of the training data set is large.

## I. INTRODUCTION

The number of road traffic deaths, ranked the eighth leading cause of death, continues to rise steadily, reaching 1.35 million in 2016 [1]. 90% of vehicle road traffic accidents are related to human factors, 41% of which is caused by human errors in identifying the dangerous condition [2]. Using crash risk prediction models to identify crash-prone traffic conditions has been regarded as a pro-active safety solution to help drivers from dangerous condition [3]. Additionally, some advanced proactive crash risk prediction models have showed promising effects on reducing the crash occurrence along with the variable speed limit system [4] and collision warning system [5], etc.

In previous studies, based on the comparison of traffic patterns between the crash occurrence and randomly selected non-crash cases, both traditional statistical models and machine learning algorithms have been employed to identify the "crash precursor condition". Specially, popular traditional statistical models such as matched case-control logistic regression [6][7], and another five advanced machine learning modeling techniques, namely artificial neural network models [8], stochastic gradient boosting technique [9], support vector machine [3], Bayesian belief networks [10], and dynamic Bayesian network [11] have been shown to be reasonable for fitting the historical crash and traffic data. Although these real-time crash risk prediction models have been proven to be capable of differentiating between crash and non-crash cases, these models have some limitations, because of the lack of consideration on the importance of the influential factors.

To solve these problems, the random forests method was employed to identify and rank the importance of variables leading to fog-related crashes on freeways [12]. And then, a Bayesian logistic regression model was employed to develop the real-time crash risk prediction model. Due to the limitation of a linear relationship assumption in Bayesian logistic regression models, this paper proposed import vector machines (IVMs) and support vector machines (SVMs), as a real-time crash risk prediction model, to identify the crash-prone traffic conditions based on the most important factors. Furthermore, compared with SVM models, IVM models can not only perform as well as SVM models in binary classification, but also can naturally be generalized to the multi-class case. Hence, this model could be more promising in identifying specific dangerous traffic conditions, such as crash severity.

## II. DATA PREPARATION AND PROCESSING

The datasets used in this study are: a) crash data from April 2014, obtained from the Shanghai Urban Expressway System in China and b) traffic data recorded by dual Loops Detectors (LDs) along the mainline of urban expressways (see details in [13]). More specifically, data were collected on Yan'an elevated road, Yixian elevated road, North-South elevated road. The LDs dataset provides the 20-second raw loop detector data for each lane, including vehicle count, vehicle speed, and time occupancy.

Similar to previous studies (e.g., [3][14][15][16]), 5-min

*Research supported by the European Union's Horizon 2020 research and innovation programme i–DREAMS under grant agreement No 814761.

K. Yang is with the Department of Civil, Geo and Environmental Engineering, Technical University of Munich, Munich, Germany (corresponding author to provide e-mail: kui.yang@tum.de).

W. Zhao is with School of Traffic and Transportation Engineering, Central South University, Changsha, China (e-mail: stacy_zhao@126.com).

C. Antoniou is with the Department of Civil, Geo and Environmental Engineering, Technical University of Munich, Munich, Germany (e-mail: c.antoniou@tum.de).

traffic data ending at five minutes prior to crash occurrences were collected to clearly identify traffic phase transitions. For instance, if a crash occurred at 14:00, the traffic data were extracted from 13:50 to 13:55. The purpose of omitting the data from 13:55 to 14:00 was to compensate for any inaccuracies in the reported crash occurrence time (e.g., [8][13][17]) and to identify hazardous traffic conditions ahead of the time of the crash occurrence to make preemptive measures possible [3][18]. A matched case-controlled structure, in which the crash cases are the traffic data before crash occurrences, was used in our studies, like previous studies (e.g., [3][14][15]). Previous studies suggested that the statistical ability is negligible when using a control-to-case ratio beyond 4:1 (e.g., [17][19][20]). Therefore, the control-to-case ratio of 4:1 was used. For each crash case, four paired observations of the non-crash traffic data were randomly selected on the basis of two matching factors, including the time of the day and the location. For example, for a crash that occurred at 15:25, traffic data taken at the detector station on the same roadway segment from 15:15 to 15:20 were included in the crash cases as an observation. Then the paired crash-free traffic data taken at the same loop detector station during the same period on four randomly selected crash-free days were used as four observations in the non-crash cases.

The extracted 20-s raw detector data for crash and non-crash cases were further aggregated into 5-min intervals so as to reduce the random noise (e.g., [8][13]) and build the mean, standard deviation and coefficient of variability (CV) of speed, volume, and occupancy, presented in TABLE I. Note that three roadway segments nearest to the crash segment were identified, namely upstream segment (U), crash segment (C) and downstream segment. Finally, the dataset has 524 crash and 2,090 non-crash scenarios.

TABLE I. DESCRIPTION OF VARIABLES

| Segment | Variable | Description |
|---|---|---|
| U | Mean_Flow_U | Mean of flow for U segment during 5-min period (pcu/lane/20s) |
| | Std_Flow_U | Std Dev of flow for U segment during 5-min period (pcu/lane/20s) |
| | CV_Flow_U | Coefficient of variability of flow for U segment during 5-min period (pcu/lane/20s) |
| | Mean_Speed_U | Mean of speed for U segment during 5-min period (km/h) |
| | Std_Speed_U | Std Dev of speed for U segment during 5-min period (km/h) |
| | CV_Speed_U | Coefficient of variability of speed for U segment during 5-min period (km/h) |
| | Mean_Occupancy_U | Mean of occupancy for U segment during 5-min period (%) |
| | Std_Occupancy_U | Std Dev of occupancy for U segment during 5-min period (%) |
| | CV_Occupancy_U | Coefficient of variability of occupancy for U segment during 5-min period (%) |
| C | Mean_Flow_C | Mean of flow for C segment during 5-min period (pcu/lane/20s) |
| | Std_Flow_C | Std Dev of flow for C segment during 5-min period (pcu/lane/20s) |
| | CV_Flow_C | Coefficient of variability of flow for C segment during 5-min period (pcu/lane/20s) |
| | Mean_Speed_C | Mean of speed for C segment during 5-min period (km/h) |
| | Std_Speed_C | Std Dev of speed for C segment during 5-min period (km/h) |
| | CV_Speed_C | Coefficient of variability of speed for C segment during 5-min period (km/h) |
| | Mean_Occupancy_C | Mean of occupancy for C segment during 5-min period (%) |
| | Std_Occupancy_C | Std Dev of occupancy for C segment during 5-min period (%) |
| | CV_Occupancy_C | Coefficient of variability of occupancy for C segment during 5-min period (%) |
| D | Mean_Flow_D | Mean of flow for D segment during 5-min period (pcu/lane/20s) |
| | Std_Flow_D | Std Dev of flow for D segment during 5-min period (pcu/lane/20s) |
| | CV_Flow_D | Coefficient of variability of flow for D segment during 5-min period (pcu/lane/20s) |
| | Mean_Speed_D | Mean of speed for D segment during 5-min period (km/h) |
| | Std_Speed_D | Std Dev of speed for D segment during 5-min period (km/h) |
| | CV_Speed_D | Coefficient of variability of speed for D segment during 5-min period (km/h) |
| | Mean_Occupancy_D | Mean of occupancy for D segment during 5-min period (%) |
| | Std_Occupancy_D | Std Dev of occupancy for D segment during 5-min period (%) |
| | CV_Occupancy_D | Coefficient of variability of occupancy for D segment during 5-min period (%) |

III. METHODOLOGY

In this section, three major methodologies used in this study were described: (i) random forest model, (ii) support vector machine model, and (iii) import vector machine model. The random forest model was used to identify the important variables, while the import vector machine model and support vector machine model were employed to predict crash risk.

*A. Random Forest for Variable Selection*

Random forest (RF) has been frequently used to identify the variable importance in traffic safety analysis (e.g., [18][21][22]). A random forest is a classifier consisting of a collection of tree-structured classifiers $\{h(x, \ominus_k), k = 1, \cdots\}$, where the $\{\ominus_k\}$ are independent and identically distributed (IID) random vectors and each tree casts a unit vote for the most popular class at input $x$ [23]. RF has the ability to solve the multi-collinearity problem of candidate variables. It has demonstrated high capability in obtaining unbiased and stable results, without a need for a separate cross-validation test data set [24]. Furthermore, RF runs efficiently on large datasets, even with thousands of variables.

Therefore, the RF technique was used in this study to identify the important traffic variables that significantly affected the crash risks. RF variable importance of $X^j$ is defined as follows [25]. For each tree $t$ of the forest, consider the associated $OOB_t$ (out-of-bag, OOB) sample (data not included in the bootstrap sample used to construct $t$). Denote by $errOOB_t$ the error (MSE for misclassification rate for classification) of a single tree $t$ on this $OOB_t$ sample. Now, randomly permute the values of $X^j$ in $OOB_t$ to get a perturbed sample denoted by $\widetilde{OOB}_t^j$ and compute $err\widetilde{OOB}_t^j$, the error of predictor $t$ on the perturbed sample. Variable importance of $X^j$ is then equal to:

$$VI(X^j) = \frac{1}{ntree}\sum_t(err\widetilde{OOB}_t^j - errOOB_t) \qquad (1)$$

where the sum is over all trees $t$ of the RF and $ntree$ denotes the number of trees of the RF.

*B. Support Vector Machine*

In standard classification problems, we are given a set of training data $\{(x_1, y_1), (x_2, y_2), \cdots, (x_N, y_N)\}$, where the output $y_i$ is qualitative and assumes values in a finite set $C$. We wish to find a classification rule from the training data, so that when given a new input $x$, we can assign a class $c$ from $C$ to it. Usually it is assumed that the training data are an independently and identically distributed (IID) sample from an unknown probability distribution $P(X, Y)$.

The standard SVM produces a non-linear classification boundary in the original input space by constructing a linear boundary in a transformed version of the original input space. The dimension of the transformed space can be very large, even infinite in some cases. This seemingly prohibitive computation is achieved through a positive definite reproducing kernel $K$, which gives the inner product in the transformed space [26].

Fitting an SVM is equivalent to minimizing [26]:

$$\frac{1}{N}\sum_{i=1}^N(1 - y_i f(x_i))_+ + \lambda\|f\|_{H_K}^2 \qquad (2)$$

with $f = b + h, h \in H_K, b \in R$. $H_K$ is the Reproducing Kernel Hilbert Space (RKHS) generated by the kernel $K$. The classification rule is given by $sign[f]$. By the representer theorem [27], the optimal $f(x)$ has the form:

$$f(x) = b + \sum_{i=1}^N a_i K(x, x_i) \qquad (3)$$

where $K(x, x_i)$ is $\emptyset(x_i)^T \emptyset(x_j)$ is the kernel function.

In this study, the Radial-basis function (RBF) kernel and linear kernel were considered:

Linear kernel: $K(x_i, x_j) = x_i^T x_j$

Radial-basis function kernel: $K(x_i, x_j) = exp(-\lambda|x_i - x_j|^2)$

*C. Import Vector Machine*

Following the tradition of logistic regression, we let $y_i \in \{0,1\}$. For notational simplicity, the constant term in the fitted function is ignored.

In the kernel logistic regression, we want to minimize the regularized negative log-likelihood [26]:

$$H = -\sum_{i=1}^N[y_i f(x_i) - ln(1 + exp(f(x_i)))] + \frac{\lambda}{2}\|f\|_{H_K}^2$$

From Eq. (2), it can be shown that this is equivalent to the finite dimensional form:

$$H = -\vec{y}^T(K_a\vec{a}) + \vec{1}^T ln(1 + exp(K_a\vec{a})) + \frac{\lambda}{2}\vec{a}^T K_q \vec{a} \qquad (4)$$

where $\vec{a} = (a_1, a_2, \cdots, a_N)^T$; the regressor matrix $K_a = [K(x_i, x_j)]_{N\times N}$; and the regularization matrix $K_q = K_a$

To find $\vec{a}$, we set the derivative of $H$ with respect to $\vec{a}$ equal to 0, and use the Newton-Raphson method to iteratively solve the score equation. It can be shown that the Newton-Raphson step is a weighted least squares step:

$$\vec{a}^{(k)} = (K_a^T W K_a + \lambda K_q)^{-1} K_a^T W \vec{z} \qquad (5)$$

where $\vec{a}^{(k)}$ is the value of $\vec{a}$ in the $k$th step, $\vec{z} = (K_a\vec{a}^{(k-1)} + W^{-1}(\vec{y} - \vec{p}))$. The weight matrix is $W = diag[p(x_i)(1 - p(x_i))]_{N\times N}$.

Basic algorithm:

*(B1)* Let $S = \emptyset, R = \{x_1, x_2, \cdots, x_N\}, k = 1$

*(B2)* For each $x_l \in R$, let

$$f_l(x) = \sum_{x_j \in S \cup \{x_l\}} a_j K(x_i, x_j)$$

Find $\vec{a}$ to minimize

$$H\{x_l\} = -\sum_{i=1}^N[y_i f_l(x_i) - ln(1 + exp(f_l(x_i)))] + \frac{\lambda}{2}\|f_l\{x\}\|_{H_K}^2 = -\vec{y}^T(K_a^l\vec{a^l}) + \vec{1}^T ln(1 + exp(K_a^l\vec{a^l})) + \frac{\lambda}{2}\vec{a^l}^T K_q^l \vec{a^l} \qquad (6)$$

where the regressor matrix $K_a^l = [K(x_i, x_j)]_{N\times(q+1)}, x_i \in \{x_1, x_2, \cdots, x_N\}, x_j \in S \cup \{x_l\}$; the regularization matrix $K_q^l = [K(x_j, x_l)]_{(q+1)\times(q+1)}, x_j \in S \cup \{x_l\}; q = |S|$.

*(B3)* Let

$x_{l^*} = argmin_{x_l \in R} H(x_l)$.
Let $S = S \cup \{x_{l^*}\}, R = R\setminus\{x_{l^*}\}, H_K = H(x_{l^*}), k = k + 1$.

*(B4)* Repeat steps (B2) and (B3) until $H_K$ converges.

*D. Evaluation Criterion*

For the purpose of comparing the performance, the Area Under the Receiver Operating Characteristic curve (AUC), sensitivity and $1 - Specificity$ were used as criteria in this paper. This Receiver Operating Characteristic (ROC) curve is created by plotting the true positive rate (i.e. Sensitivity) against the false positive rate (1 - Specificity) at thresholds ranging from 0 to 1, and AUC indicates the predictive performance of model, which is between 0.5 and 1. The model with the greater value of AUC has the better performance.

sensitivity and $1 - Specificity$ can be computed as:

$$Sensitivity = \frac{TP}{TP + FN}$$
$$1 - Specificity = \frac{FP}{FP + TN}$$

where, $TP$ denotes the number of crashes predicted correctly as crashes, $FP$ denotes the number of non-crashes wrongly predicted as crashes, $FN$ denotes the number of crashes wrongly predicted as non-crashes, and finally $TN$ denotes the number of non-crashes correctly predicted as non-crashes.

## IV. ANALYSIS RESULTS

### A. Variable Selection Using RF

As shown in TABLE I. , 27 candidate variables were obtained in this paper. The RF technique was used to identify important variables that significantly affected the crash risk. The "random Forest" package [28][29] in the R software was used to develop the RF models. The number of trees in the forest and the number of variables tried at each node should be identified firstly in the RF model. OOB error rates for different numbers (i.e., from 10 to 10,000) of trees in RF models were calculated, and it was found that the OOB error rates started to be stable at a low level after 400 trees. Thus, 400 trees were used in the RF model to obtain stable estimations of variable importance. And then the RF models with different numbers of variables tried at each node were estimated and compared. The OOB error rate reached a minimum, when the number of variables tried at each node was equal to 2. Figure 1. shows the final results of variable importance rankings; the mean decrease accuracy (%IncMSE) was the selection criterion. It can be concluded from the figure that the top 6 variables are the most important factors. It is expected that most of the top variables are from the crash segment since their variables are the closet to the crash occurrence from the spatio-temporal aspect. And then, the variables from the traffic upstream are more important to the crash occurrence than those from the traffic downstream since several minutes later the traffic flow in the upstream will arrive / pass the crash segments.

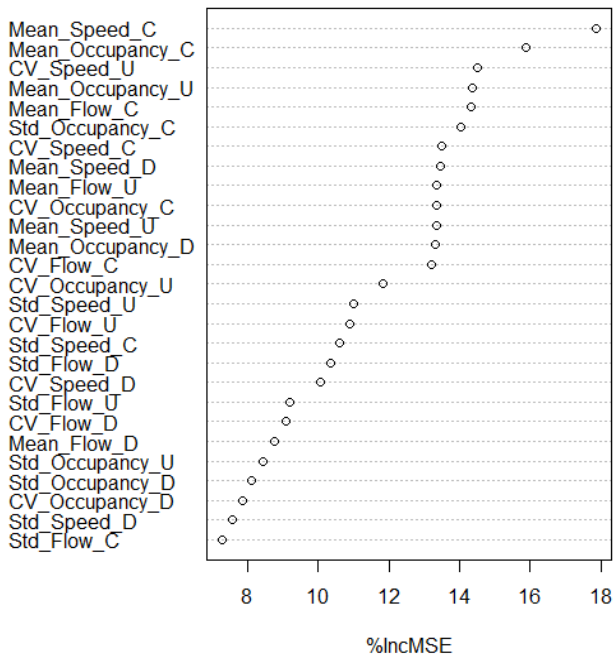

Figure 1. Variable importance provided by random forest.

Considering the correlation test results to avoid multicollinearity problems, Mean_Speed_C, CV_Speed_U, Mean_Flow_C and Std_Occupancy_C were selected as inputs in the modeling steps. TABLE II. lists summary statistics for variables for model.

TABLE II. SUMMARY STATISTICS FOR VARIABLES FOR MODELS

| Variable | Mean | Std. Dev | Min | Max |
|---|---|---|---|---|
| Mean_Speed_C | 52.40 | 21.47 | 7.59 | 94.34 |
| CV_Speed_U | 0.34 | 0.23 | 0.06 | 1.98 |
| Mean_Flow_C | 5.82 | 2.78 | 0.01 | 11.80 |
| Std_Occupancy_C | 9.66 | 7.59 | 0.13 | 44.37 |

### B. Crash Risk Prediction Model: Support Vector Machine Models

On the basis of the same matched case-control dataset, the Packages 'e1071' [29][30] in R®3.5.1 was used to develop the SVM models, varying some key parameters (the kernel function, the gamma and the cost). The four same variables were used as the inputs of SVM models so as to avoid the impact of different variables. Ten different gammas (i.e. 0.001, 0.01, 0.1, 0.5, 1, 2, 5, 10, 20, and 50) and five different costs (i.e. 0.01, 0.01, 1, 10, and 100) were considered for each of two kernel functions (i.e. radial and linear). Finally, the best model was identified for each of the kernel functions on the basis of the classification error from a total of 100 SVM models (i.e. $10 \times 5 \times 2 = 100$). TABLE III. lists the results for SVM models.

TABLE III. RESULTS OF SVM MODELS

| Kernel | Radial | Linear |
|---|---|---|
| Gamma | 1 | 0.001 |
| Cost | 1 | 0.01 |
| Number of Support Vectors | 805 | 740 |
| Best performance | 0.187 | 0.200 |

### C. Crash Risk Prediction Model: Import Vector Machine Models

The MATLAB®2017b [31] was used to develop the IVM models by varying some key parameters (e.g. regularization parameter, and kernel parameter). The same training dataset was employed to develop the IVM models. 20 different kernel parameters (i.e. 1, 2, 3, … , and 20) and 20 different regularization parameter (i.e. 1, 2, 3, … , and 20) were employed to structure 100 IVM modes. According to the prediction performance, the best model was identified. The parameters of the best model are listed in TABLE IV. , whose evolution of the negative log likelihood over different iterations is shown as Figure 2. . The error of the model drops sharply and tends to be stable from the 10th iteration. Note that the number of used import vectors in IVM model is only 15, which is much smaller than the SVM models whose number of support vectors are 805 and 740. Thus, IVM model has a much smaller fraction than the SVM models, and typically its training time is much shorter than the SVM models in our experiences.

TABLE IV. RESULTS OF IVM MODELS

| Parameters | Values |
|---|---|
| Kernel Sigma: kernel parameter | 22 |
| $\lambda$: regularization parameter | 0.0067 |
| $S$: number of used import vectors | 15 |
| $a$: decision hyperplane | [15×2 double] |
| Training time | 2.3088s |

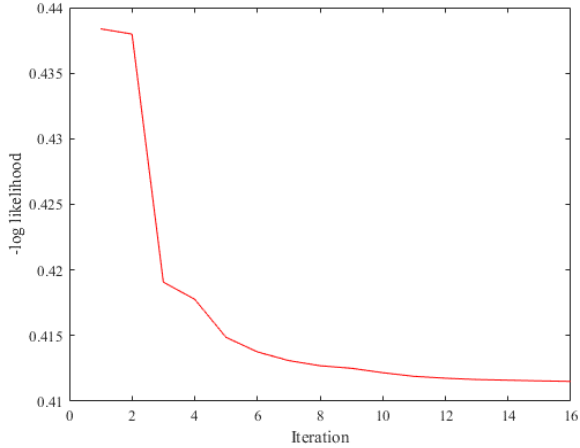

Figure 2. The evolution of the negative log likelihood over different iterations.

*D. Predictive Performance Evaluation*

The matched case-control dataset was randomly divided into training data and test data. The training data has 70% of matched case-controls in the original dataset, while test data has 30% of matched case-controls in the original dataset. The training data was employed to build the SVM models and IVM model on the basis of the parameter settings above. And the test data was used to test the models via the AUC. The results are listed as TABLE IV.

From the training data, the AUC values of IVM model and SVM model based on radial function are significantly higher than the SVM model base on linear function. However, the AUC of the SVM model based on radial function from the test data was the lowest due to the overfitting. The IVM model obtained the highest AUC (i.e. 0.809) from the test data, which indicates that the prediction performance of IVM is the best. The ROC curves of different models are shown in Figure 3. . There is no significant difference between the IVM model and the SVM model based on radial function when the 1-Specificity is less than around 30%. And there is no significant difference between the IVM model and the SVM model based on linear function when the 1-Specificity is greater than around 30%. The prediction performance is measured by the percentage of the correctly predicted crash cases for different false-alarm rates (1-specificity) in TABLE V. 40.8% of crashes can be identified correctly by the IVM model at the cost of 10.1% false-alarm rate, which makes the total accuracy reach 80.0%. And 60.5% of crashes can be identified correctly by the IVM model at the cost of 20.0% false-alarm rate, which makes the total accuracy reach 76.1%. Totally, the classification ability of IVM model is better or similar to that of SVM models. Considering IVM model has a much smaller fraction, which shortens typically its training time, IVM model therefore has greater advantages and has more potentials in applications, such as identifying the dangerous pro-active traffic conditions.

TABLE V. PREDICTIVE PERFORMANCES OF THE THREE MODELS

| Models | Training Data | Test Data | | | |
|---|---|---|---|---|---|
| | *AUC* | *AUC* | *1-Specificity* | *Sensitivity* | *Accuracy* |
| IVM | 0.835 | 0.809 | 30.1% | 73.9% | 70.7% |
| | | | 20.0% | 60.5% | 76.1% |
| | | | 10.1% | 40.8% | 80.0% |
| SVM (Kernel="Linear") | 0.788 | 0.790 | 30.1% | 75.8% | 71.1% |
| | | | 20.0% | 53.5% | 74.7% |
| | | | 10.1% | 28.7% | 77.6% |
| SVM (Kernel="Radial") | 0.839 | 0.764 | 30.1% | 71.3% | 70.2% |
| | | | 20.0% | 58.6% | 75.7% |
| | | | 10.1% | 40.1% | 79.9% |

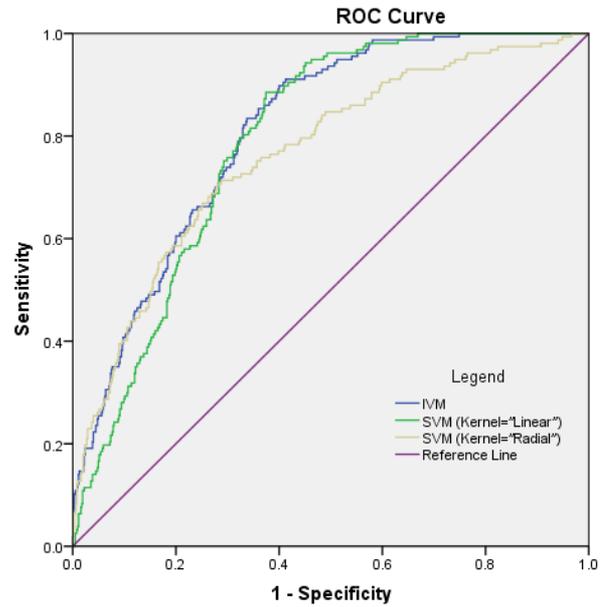

Figure 3. The ROC curves of different models.

## V. DISCUSSION

The IVM model provides a kernel-based, probabilistic and discriminative learner, and inherently contains a reconstructive component. It could be better to output the probability. Besides, IVM model could be further combined with incremental learning to handle big data [32]. The proper incremental learning scheme could deal with adding and removing data samples and the update of the current set of model parameters of IVM classifiers.

Compared with the existing real-time crash risk prediction models (crash or no-crash), IVM can not only perform as well in identifying the accident-involved conditions, but also be helpful in identifying multi-level risk of these conditions. Based on the identification of high-risk traffic conditions, the traffic authorities can take some pro-active measures to help drivers to avoid this dangerous conditions, such as providing roadside guidance signs to drivers. Besides, based on the speed difference between upstream and downstream vehicles, IVM can classify the traffic conditions into several levels as the thresholds to provide real-time speed limits by the variable speed limit system, thus help drivers to avoid being involved in accidents. Furthermore, the risky condition of roads should be incorporated into route choice planning, which thus helps drivers to achieve high-frequency accidents and high-risk level roads aversions.

REFERENCES

[1] World Health Organization. Global status report on road safety 2016. World Health Organization, 2018

[2] NHTSA. Traffic Safety Facts: Crash Status – A Brief Statistical Summary. Report No. DOT-HS-812-115, U.S. Department of Transportation, Washington, D.C., 2015.

[3] Yu, R. and Abdel-Aty, M., (2013). Utilizing support vector machine in real-time crash risk evaluation. Accident Analysis & Prevention, 51, pp.252-259.

[4] Lee C, Abdel-Aty M., (2008). Testing effects of warning messages and variable speed limits on driver behavior using driving simulator[J]. Transportation Research Record, 2069(1): 55-64.

[5] Cicchino J B., (2017). Effectiveness of forward collision warning and autonomous emergency braking systems in reducing front-to-rear crash rates. Accident; analysis and prevention, 99(Pt A):142-152.

[6] Xu, C., Liu, P., Wang, W., (2016). Evaluation of the predictability of real-time crash risk models. Accident Analysis & Prevention 94, 207-215.

[7] Xu, C., Liu, P., Wang, W., Li, Z., (2012). Evaluation of the impacts of traffic states on crash risks on freeways. Accident Analysis & Prevention 47, 162-171.

[8] Pande, A., Das, A., Abdel-Aty, M., Hassan, H., (2011). Real-time crash risk estimation: Are all freeways created equal? In: Compendium of Papers CD-ROM, Transportation Research Board 2011 Annual Meeting. Washington, DC.

[9] Ahmed, M. M., Abdel-Aty, M., (2013). Application of stochastic gradient boosting technique to enhance reliability of real-time risk assessment: use of automatic vehicle identification and remote traffic microwave sensor data. Transportation Research Record 2386, 26-34.

[10] Hossain, M., Muromachi, Y., (2012). A Bayesian network based framework for real-time crash prediction on the basic freeway segments of urban expressways. Accident Analysis & Prevention 45, 373-381.

[11] Sun, J., Sun, J, (2015). A dynamic Bayesian network model for real-time crash prediction using traffic speed conditions data. Transportation Research Part C: Emerging Technologies 54, 176-186.

[12] Zhai B, Lu J, Wang Y, et al., (2020). Real-time prediction of crash risk on freeways under fog conditions. International Journal of Transportation Science and Technology.

[13] Yang, K., Yu, R., Wang, X., Quddus, M., & Xue, L. (2018). How to determine an optimal threshold to classify real-time crash-prone traffic conditions?. Accident Analysis & Prevention, 117, 250-261.

[14] Xu, C., Wang, W., & Liu, P. (2013). Identifying crash-prone traffic conditions under different weather on freeways. Journal of safety research, 46, 135-144.

[15] Xu, C., Tarko, A. P., Wang, W., & Liu, P. (2013). Predicting crash likelihood and severity on freeways with real-time loop detector data. Accident Analysis & Prevention, 57, 30-39.

[16] Yang, K., Wang, X., & Yu, R. (2018). A Bayesian dynamic updating approach for urban expressway real-time crash risk evaluation. Transportation Research Part C: Emerging Technologies, 96, 192-207.

[17] Xu, C., Liu, P., Wang, W., & Li, Z. (2015). Safety performance of traffic phases and phase transitions in three phase traffic theory. Accident Analysis & Prevention, 85, 45-57.

[18] Xu, C., Wang, W., & Liu, P. (2013). A Genetic Programming Model for Real-Time Crash Prediction on Freeways. IEEE Trans. Intelligent Transportation Systems, 14(2), 574-586.

[19] Ahrens, W., Pigeot, I., (2005). Handbook of Epidemiology. Springer.

[20] Rothman, K.J., Greenland, S., (1998). Modern Epidemiology, second ed. LippincottWilliams and Wilkins, Philadelphia.

[21] Ahmed, M. M., & Abdel-Aty, M. A. (2012). The viability of using automatic vehicle identification data for real-time crash prediction. IEEE Transactions on Intelligent Transportation Systems, 13(2), 459-468.

[22] Yu, R., & Abdel-Aty, M. (2014). Analyzing crash injury severity for a mountainous freeway incorporating real-time traffic and weather data. Safety science, 63, 50-56.

[23] Breiman, L. (2001). Random forests. Machine learning, 45(1), 5-32.

[24] Xu, C., Liu, P., Wang, W., & Li, Z. (2014). Identification of freeway crash-prone traffic conditions for traffic flow at different levels of service. Transportation research part A: policy and practice, 69, 58-70.

[25] Genuer, R., Poggi, J. M., & Tuleau-Malot, C. (2010). Variable selection using random forests. Pattern Recognition Letters, 31(14), 2225-2236.

[26] Zhu, J., & Hastie, T. (2002). Kernel logistic regression and the import vector machine. In Advances in neural information processing systems (pp. 1081-1088).

[27] Kimeldorf, G., & Wahba, G. (1971). Some results on Tchebycheffian spline functions. Journal of mathematical analysis and applications, 33(1), 82-95.

[28] Leo Breiman, Adele Cutler, Andy Liaw and Matthew Wiener (2018). randomForest: Breiman and Cutler's random forests for classification and regression. https://cran.r-project.org/web/packages/randomForest/randomForest.pdf

[29] R Core Team (2020). R: A language and environment for statistical computing. R Foundation for Statistical Computing, Vienna, Austria.

[30] Meyer, D., Dimitriadou, E., Hornik, K., Weingessel, A., Leisch, F. e1071: Misc functions of the department of statistics, probability theory group (Formerly: E1071), TU Wien; [R Package e1071 Version 1.6-8], 2017. < https://CRAN.R-project.org/package=e1071>

[31] MathWorks, Inc. (2017). MATLAB 2017b.

[32] Roscher, R., Förstner, W., & Waske, B. (2012). I2VM: Incremental import vector machines. Image and Vision Computing, 30(4-5), 263-278.